\documentclass{article}

\usepackage[preprint]{neurips_2021}

\usepackage[utf8]{inputenc} 
\usepackage[T1]{fontenc}    
\usepackage{hyperref}       
\usepackage{url}            
\usepackage{booktabs}       
\usepackage{amsfonts}       
\usepackage{nicefrac}       
\usepackage{microtype}      
\usepackage{xcolor}         

\usepackage{graphicx}
\usepackage{bm}
\usepackage{subcaption}

\title{On the Efficacy of Co-Attention Transformer Layers in Visual Question Answering}

\author{%
  Ankur Sikarwar \\
  Department of ECE \\
  Birla Institute of Technology, Mesra \\
  \texttt{ankursikarwardc@gmail.com} \\
  \And
  Gabriel Kreiman \\
  Center for Brains, Minds and Machines \\
  Harvard Medical School \\
  \texttt{gabriel.kreiman@tch.harvard.edu} \\
}

\begin{document}

\maketitle

\begin{abstract}
    In recent years, multi-modal transformers have shown significant progress in Vision-Language tasks, such as Visual Question Answering (VQA), outperforming previous architectures by a considerable margin. This improvement in VQA is often attributed to the rich interactions between vision and language streams. In this work, we investigate the efficacy of co-attention transformer layers in helping the network focus on relevant regions while answering the question. We generate visual attention maps using the question-conditioned image attention scores in these co-attention layers. We evaluate the effect of the following critical components on visual attention of a state-of-the-art VQA model: (i) number of object region proposals, (ii) question part of speech (POS) tags, (iii) question semantics, (iv) number of co-attention layers, and (v) answer accuracy. We compare the neural network attention maps against human attention maps both qualitatively and quantitatively. Our findings indicate that co-attention transformer modules are crucial in attending to relevant regions of the image given a question. Importantly, we observe that the semantic meaning of the question is \emph{not} what drives visual attention, but specific keywords in the question do. Our work sheds light on the function and interpretation of co-attention transformer layers, highlights gaps in current networks, and can guide the development of future VQA models and networks that simultaneously process visual and language streams.
\end{abstract}

\section{Introduction}

The ability of humans to efficiently ground information across different modalities, such as vision and language, plays a central role in cognitive function. The interactions between vision and language are highlighted in visual question answering (VQA) tasks, where attentional allocation is naturally routed by combination of sensory and semantic cues. For instance, given an image of people playing football and the question 'What color shirt is the person behind the referee wearing?', subjects rapidly identify the referee, saccade to the player behind the referee, and process the relevant regions of the image to find the answer. A four-year old can easily answer such questions and seamlessly direct visual attention to the relevant regions based on the question.

In contrast, such multi-modal tasks are quite challenging for current AI systems because the solution  encompasses several increasingly complex subtasks. First of all, the system has to interpret the key elements in the question for attention allocation, in this case, referees, players, and shirt. Distinguishing the referee from the players is complicated in itself, as it requires further background knowledge about sports. Next, the system has to make sense of prepositions like 'behind' to capture spatial relationships between objects or agents, in this case, to attend to one specific player. Finally, the system needs to visually attend to the task-relevant regions, distill the type of information required (shirt color), and produce the answer.

\begin{figure}[t]
  \centering
  \includegraphics[scale=0.9]{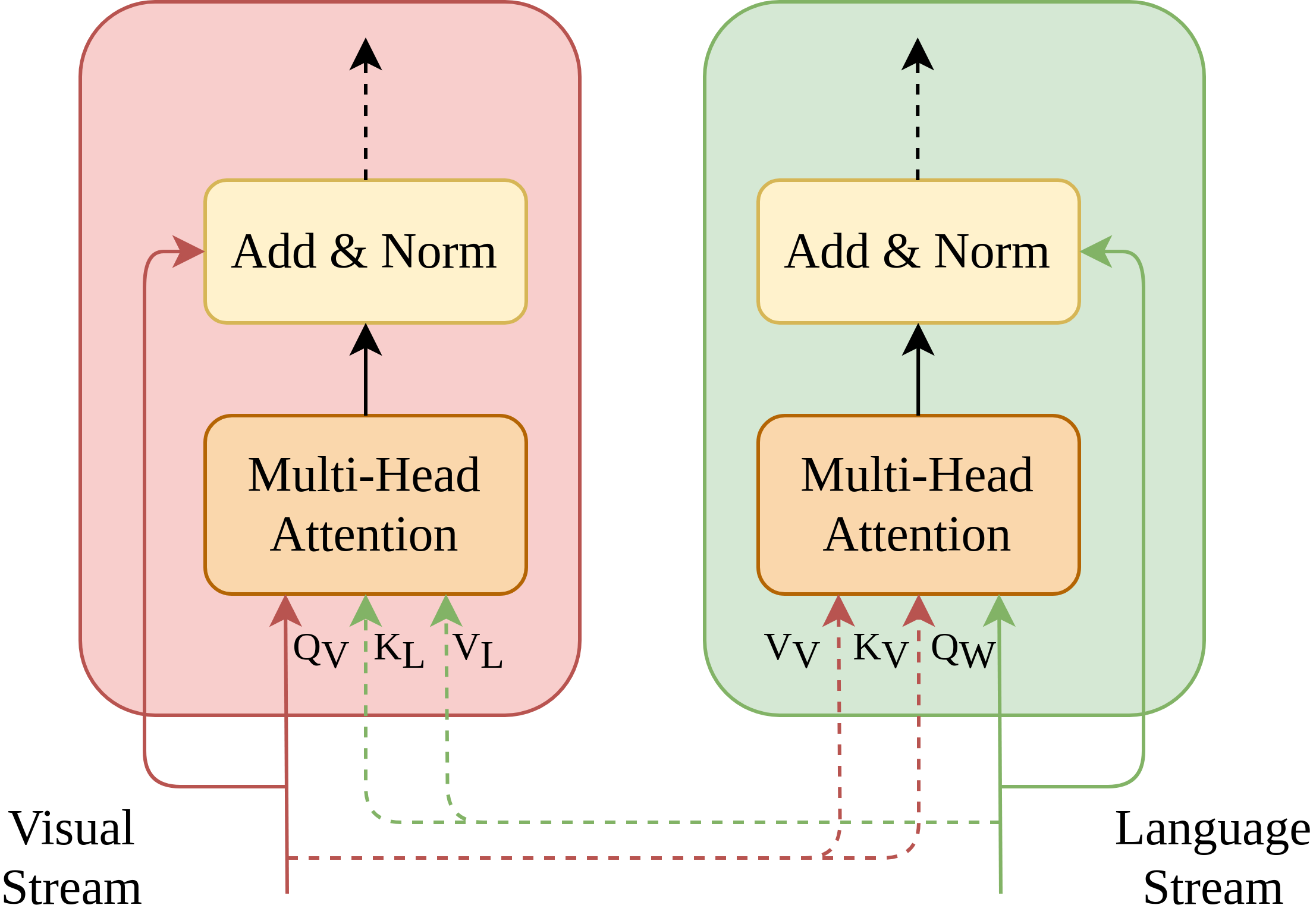}
  \caption{\textbf{Co-attention transformer layer \citep{lu2019vilbert}}}
\label{coAttentionLayer}
\end{figure}

Recently, there has been an exciting trend of extending the successful transformer architecture \citep{vaswani2017attention} to solve multi-modal tasks combining modalities including text, audio, images, and videos \citep{chuang2019speechbert,gabeur2020multi,sun2019videobert}. This trend has led to significant improvements in state-of-the-art models for Vision-Language tasks like visual grounding, referring expressions, and visual question answering. These families of models are based on either single-stream or two-stream architectures. The former shares the parameters across both modalities, while the latter has separate processing stacks for vision and language. In \citep{lu2019vilbert}, Co-Attention Transformer Layers (\textbf{Fig.~\ref{coAttentionLayer}}) are used to facilitate interactions between the visual and language streams of the network. 
The task-relevant representations from the language stream modulate processing in the visual stream in the form of attention.

In this work, we assess the capabilities of co-attention transformer layers in guiding visual attention to task-relevant regions. We focus specifically on the Visual Question Answering task and conduct experiments to gain insight into the attention mechanisms of these layers and compare these mechanisms to human attention. 
Given an image/question pair, we generate attention maps for different co-attention layers based on the question-conditioned image attention scores and evaluate these maps against human attention maps, quantitatively and qualitatively, via rank-correlation and visualizations. We ask the following questions:
    1) Does the use of object-based region proposals act as a bottleneck?
    2) Is the model more likely to correctly answer a question when its attention map is better correlated to humans?
    3) What is the role of question semantics in driving the model's visual attention?
    4) What is the importance of different parts of speech in guiding the model to attend to task-relevant regions?
Our experiments demonstrate that object-based region proposals often restrict the model from focusing on task-relevant regions. We show that rank-correlation between human and machine attention is considerably higher in current state-of-the-art transformer-based architectures compared to previous CNN/LSTM networks. Lastly, we find that question semantics have little influence on the model's visual attention, and only specific keywords in the question are responsible for driving attention.


\section{Related Work}
\label{related_work}

The Visual Question Answering (VQA) v1 dataset containing images from the MSCOCO dataset \citep{lin2014microsoft} with over 760K questions and 10M answers was introduced in \citep{antol2015vqa}, and a more balanced VQA v2 dataset was introduced in \citep{goyal2017making}. The initial model for VQA \citep{antol2015vqa} employed deep convolutional neural networks and recurrent neural networks to compute image and question representations separately. These were then fused using point-wise multiplication and fed to a Multi-Layer Perceptron (MLP) to predict the answer. Later, \citep{yang2016stacked} proposed Stacked Attention Networks (SAN), in which the question representation from an LSTM was used for predicting an attention distribution over different parts of the image. Based on this attention and the question representation, another level of attention was performed over the image. The Hierarchical Co-Attention Model \citep{lu2016hierarchical} introduced co-attention, where the model attends to parts of the image along with parts of the question. Given a question about an image, this model hierarchically uses word-level, phrase-level, and question-level co-attention.

The VQA-HAT dataset consisting of human attention maps for question/image pairs from the VQA v1 dataset was introduced in \citep{vqahat}. These maps were collected by asking humans to deblur different image regions by clicking on those regions to answer the question. Attention-based VQA models \citep{yang2016stacked, lu2016hierarchical} based on convolutional neural networks and LSTM modules, but not transformer-based models, were compared against human attention maps \citep{vqahat}. The authors concluded that these models did \emph{not} attend to the same regions as humans while answering the question. However, increased performance was weakly associated with a better correlation between human and model attention maps. 
Later, \citep{goyal2016towards} used guided backpropagation and occlusion techniques to generate image importance maps for a VQA model and then compared those with human attention maps.


Various transformer-based VQA models \citep{li2020oscar, chen2020uniter, su2019vl, li2019visualbert, li2019unicodervl, zhou2019unified, chefer2021generic} have been introduced in the last few years. Among them, \citep{tan2019lxmert} and \citep{lu2019vilbert} are two-stream transformer architectures that use cross-attention layers and co-attention layers, respectively, to allow information exchange across modalities. There are several studies on the interpretability of VQA models \citep{goyal2016towards, agrawal2016analyzing, kafle2017analysis, jabri2016revisiting}, and yet very few have focused on the co-attention transformer layers used in recent VQA models. 
In this work, we use ViLBERT \citep{lu2019vilbert} for our study as it employs these co-attention layers.



\section{Methods}
\label{method}

We study the co-attention module between language and vision and the interactions within this module. To study co-attention in two-stream vision-language transformer architectures, we evaluated visual attention in the model by comparing it against human attention maps. ViLBERT \citep{lu2019vilbert} is an extension of the BERT architecture \citep{devlin2018bert} to process visual inputs. Given a question and an image, the model processes them separately in the language and visual streams, respectively. Both visual and language streams contain a stack of transformer and co-attention transformer layers. The embeddings for the word tokens and other special tokens are fed to the language stream after adding positional embeddings. The image is processed through the Faster RCNN network \citep{ren2016faster} to generate features for different region proposals. The feature representations of region proposals with the highest objectness score are fed to the visual stream. The model then processes these inputs through the two streams while fusing information within them using subsequent \emph{co-attention layers} (\textbf{Fig.~\ref{coAttentionLayer}}).

\subsection{Setup}
\label{setup}
The ViLBERT \citep{lu2019vilbert, Lu_2020_CVPR} network variant in our study uses the BERT\textsubscript{BASE} model \citep{devlin2018bert} for the language part, composed of $12$ transformer blocks. The latter $6$ blocks have co-attention transformer modules stacked between them. The visual stream comprises $6$ transformer and co-attention transformer modules. The co-attention transformer layer uses $8$ parallel attention heads. All experiments were performed on a single NVIDIA 1080 Ti GPU. The source code will be publicly available upon publication.

\subsection{Attention Map Generation}
\begin{figure}[t]
  \centering
  \includegraphics[width=13.9cm, height=6.8cm]{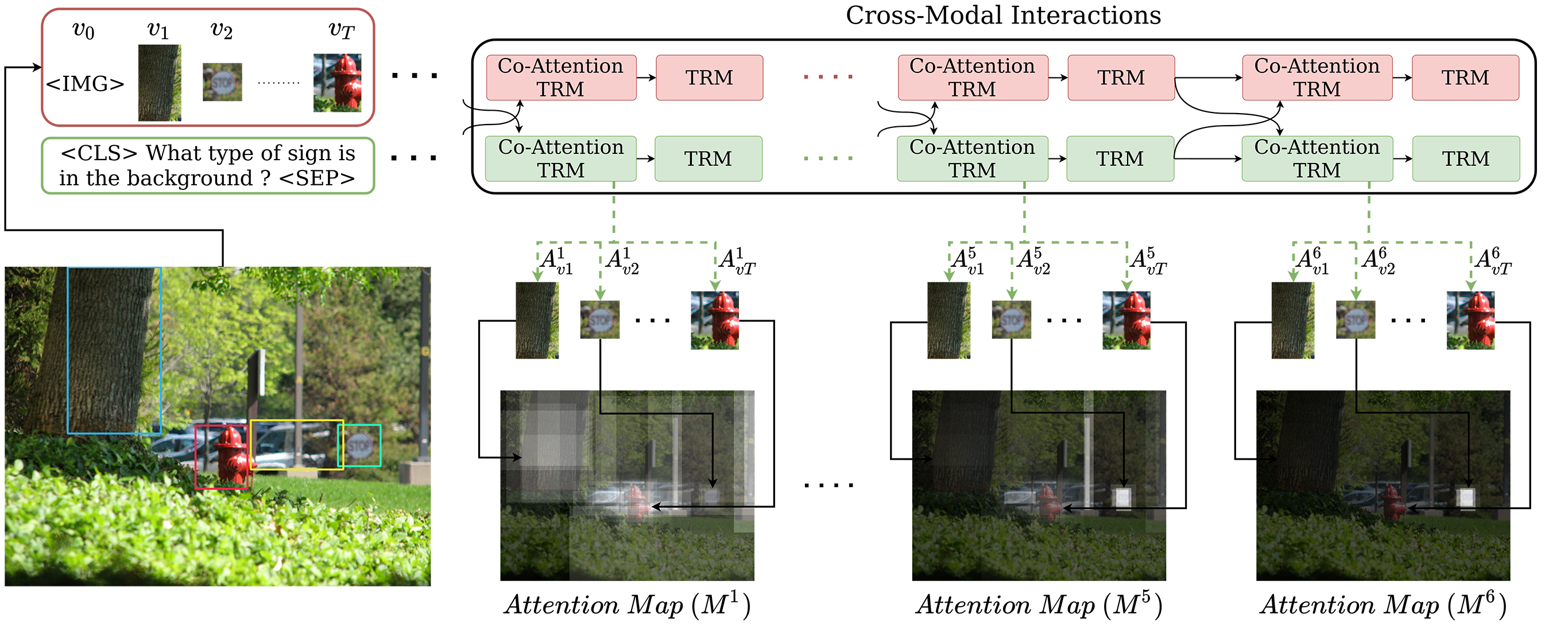}
  \caption{\textbf{Illustration of our attention map generation process.}}
  \label{attention_generation_illustration}
\end{figure}

Given an image and a question, the inputs to the visual stream are the region features $v_0, v_1, \ldots, v_T$ and the input to the language stream are $w_0, w_1, \ldots, w_N$. We generate an attention map for each co-attention transformer layer in the model as shown in \textbf{Fig. \ref{attention_generation_illustration}}. Inside the multi-head attention block in each co-attention transformer layer, the key and value matrices from one stream are projected onto another stream and vice versa. Consequently, inside the language stream, the multiplication of the Query matrix ($Q_L$) from the language stream and the Key matrix ($K_V$) from the visual stream produces attention scores over the different image regions based on the question. These attention scores are then passed through a softmax operation to generate respective attention probabilities

\[a^i_h = softmax(\frac{Q_LK^T_V}{\sqrt{d_k}}),\]

where $i$ is the co-attention layer number, $h$ is the attention head number, and $\sqrt{d_k}$ is a scaling factor \citep{vaswani2017attention}. These probabilities over the $8$ attention heads capture the modulations from each text token to different image regions. To generate question-level attention maps, we first average these attention probabilities (before dropout) over all the attention heads and then across the words present in the question. This gives us attention data $\bm{A^1}, \ldots, \bm{A^6}$ for the $6$ co-attention layers, where $\bm{A^i} = \{A^i_{v1}, \ldots, A^i_{vT}\}$. 
Based on the attention probability of different region proposal, i.e., $A^i_{v1}, \ldots, A^i_{vT}$, we weigh the corresponding pixel intensities in an image matrix and then normalize this image matrix to get the final attention map over the image, conditioned on the question. We do this for all $6$ co-attention layers to get attention maps $M^1, \ldots, M^6$. 

\subsection{Comparison Metric}
\label{comparison_metric}

We use rank-correlation (denoted by $\rho$ in the visualization figures) to compare ViLBERT's attention with human attention \citep{vqahat}. Both attention maps are scaled to $14$ x $14$ and then flattened to get a $196$ dimensional vector. These two vectors are then ranked based on their spatial attention and then we compute the correlation between the two rank vectors. All reported rank-correlation values except Question POS tag experiments (Sec. \ref{experiments_pos}), show averages over $1,374$ question/image pairs from the VQA-HAT \citep{vqahat} validation set.

\section{Experiments}
\label{experiments}

\subsection{Similarity to human attention shows a small dependence on the number of region proposals} 
\label{experiments_roi}
\begin{figure}[t]
  \centering
  \includegraphics[width=8.98cm, height=6.7cm]{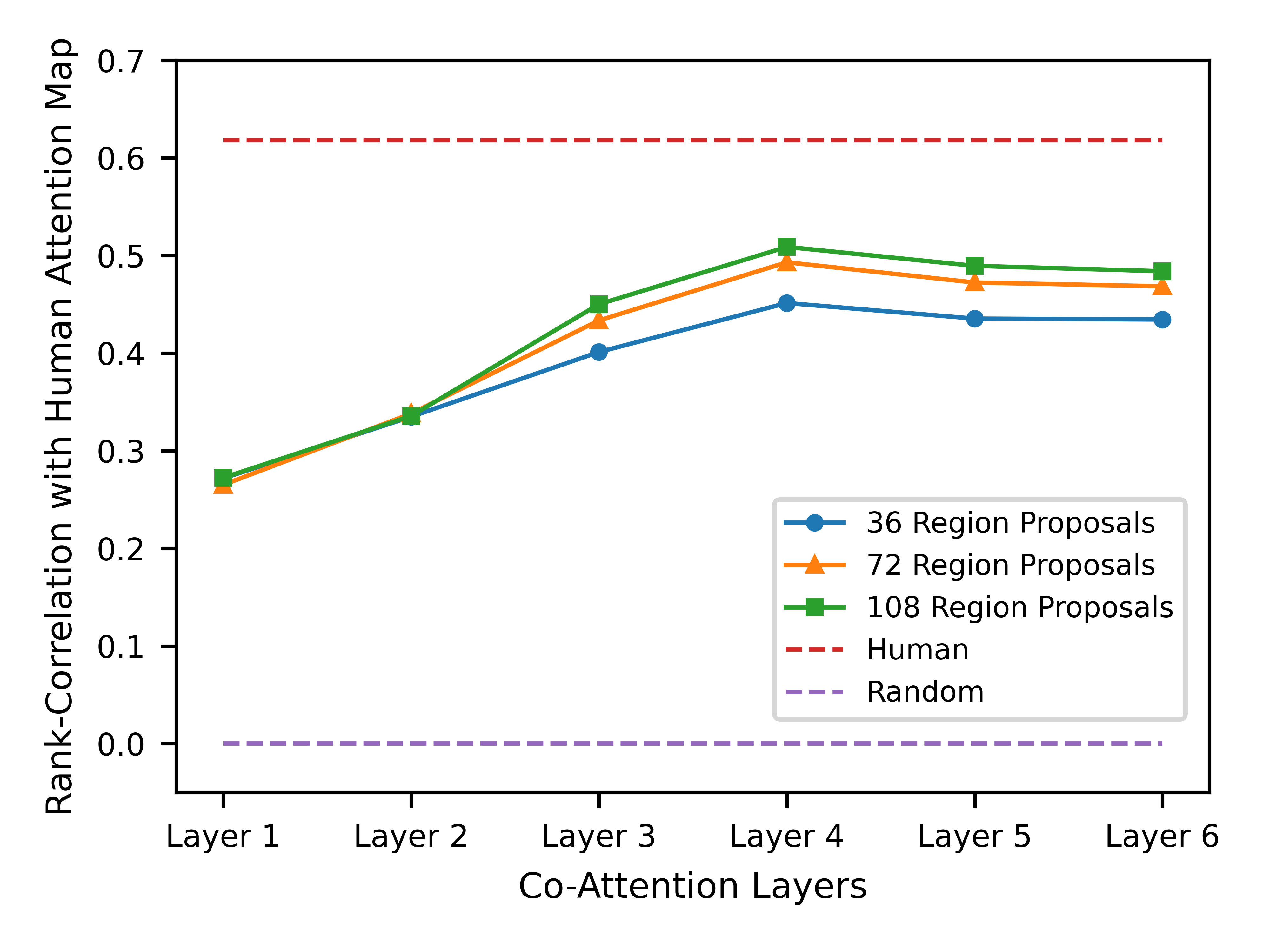}
  \caption{\textbf{The similarity between ViLBERT and human attention benefits from more region proposals.}
  The rank-correlation of ViLBERT's \citep{lu2019vilbert} attention with human attention increases monotonically up to layer 4 
   (see section~\ref{experiments_roi} for details). Error bars showing standard error of means are smaller than the symbol size in this plot.}
  \label{figure_experiments_roi}
\end{figure}

We investigated the influence of the number of region proposals on the model's ability to examine task-relevant regions. 
 Since humans rely on context to solve a problem, we hypothesize that more region proposals bring in more task-relevant context from the image, thus increasing the rank-correlation of the model's attention to that of humans and, in turn, increasing the answering accuracy. We show the rank-correlation of ViLBERT's \citep{lu2019vilbert} attention maps with human attention maps across successive co-attention layers in \textbf{Fig.~\ref{figure_experiments_roi}} for varying numbers of region proposals. To put results in perspective, we compare the results against an upper bound given by the rank-correlation for inter-human comparisons and a lower bound given by random attention allocation. 

\begin{figure}[htbp]
  \centering
  \includegraphics[scale=0.59]{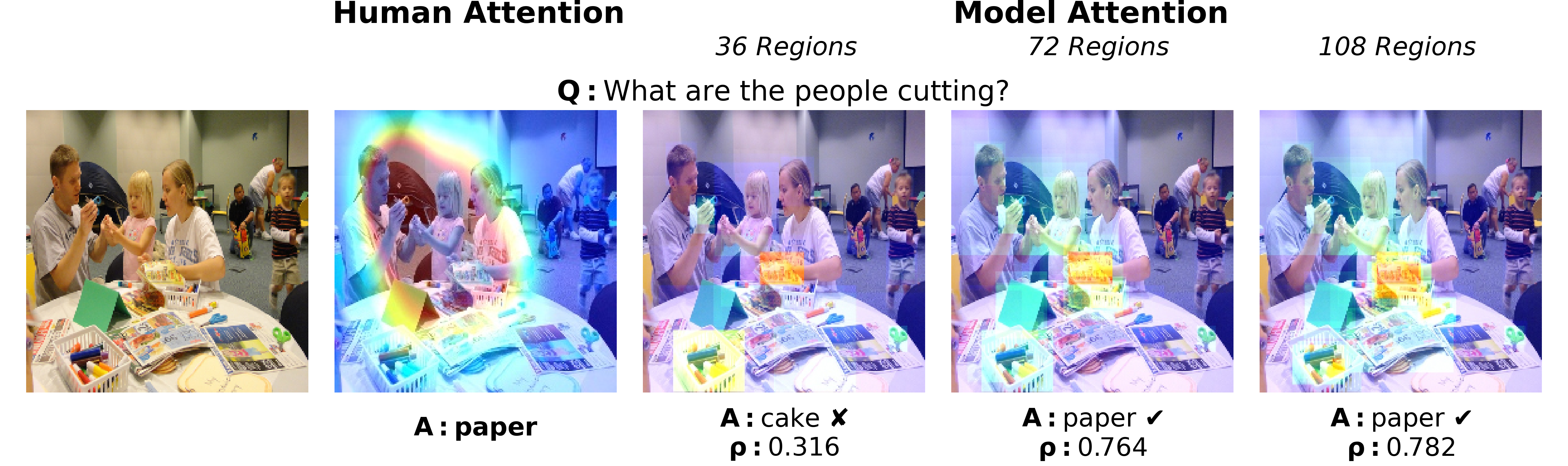}
  
  \vspace{0.08cm}
  
  \includegraphics[scale=0.59]{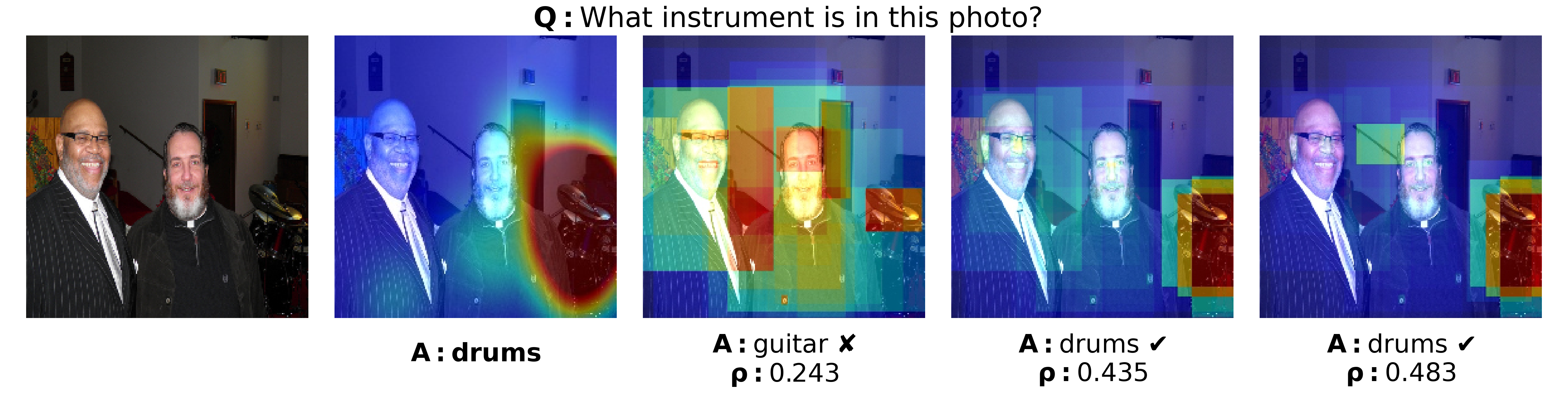}
  \includegraphics[scale=0.3]{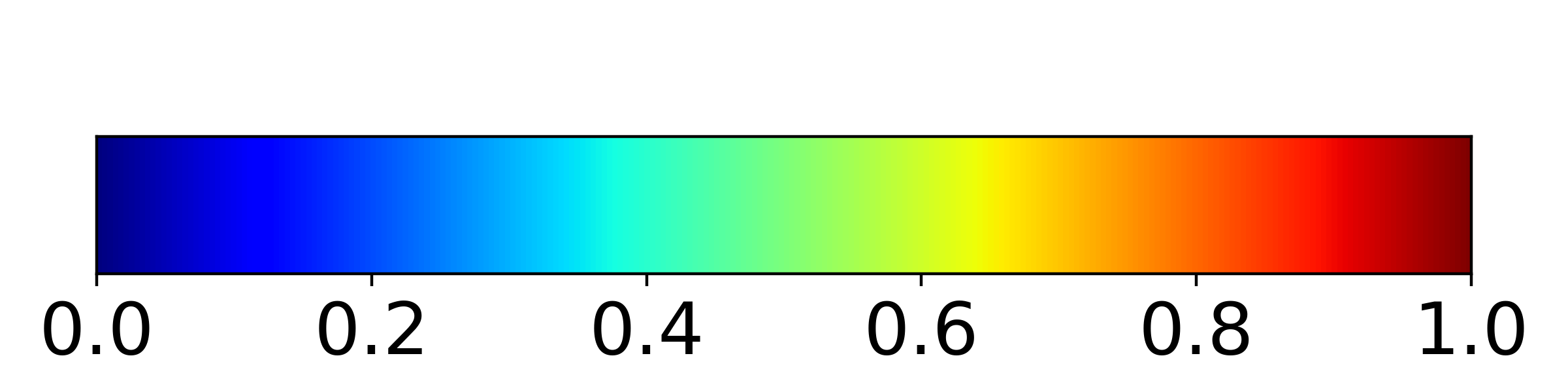}
  \caption{\textbf{Visualization for cases where increasing number of region proposals increases accuracy as well as rank-correlation with human attention.} 
  The question and answers are shown above and below the images.
  Column 1: input image, Column 2: human attention map. Columns 3, 4, 5: ViLBERT's \citep{lu2019vilbert} attention map for 36, 72, and 108 region proposals. The bottom colormap describes the intensity of the attention maps. Additional visualizations are provided in \textbf{Appendix A.1}.}
  \label{figure_roi_viz}
\end{figure}

Increasing the number of region proposals led layers 3-6 of the model to attend to regions more similar to those attended by humans. The increased context due to more region proposals also improved the model’s VQA accuracy (\textbf{Table~\ref{acc_roi_table}} and examples in \textbf{Fig.~\ref{figure_roi_viz}}). The region proposals are generated using Faster RCNN \citep{ren2016faster}, an object detection architecture. Therefore, even in the first co-attention layer, which has little interaction with the language stream, the rank-correlation of the model's visual attention with human attention is well above chance. The correlation in the lower layers is likely due to the observation that the majority of the questions in the VQA dataset \citep{antol2015vqa} focus either on object categories or object attributes that are salient in terms of basic visual features.

Given a fixed number of region proposals,  the rank-correlation increases monotonically until layer 4 and then stays approximately constant. This initial increase validates the crucial role of co-attention layers in guiding visual attention in the model. Additionally, increasing the number of region proposals captures objects' features using multiple aspect ratios and scales, often helping the model to better attend to the object in question, as depicted in the example in \textbf{Fig.~\ref{figure_roi_viz}} (row 2).

\begin{table}
  \caption{VQA accuracy of ViLBERT \citep{lu2019vilbert} with different number of region proposals. Accuracies are computed over all the question/image pairs in the VQA-HAT \citep{vqahat} validation set.\\}
  \label{acc_roi_table}
  \centering
  \begin{tabular}{ccc}
    \toprule
    Method     & VQA Accuracy      \\
    \midrule
    ViLBERT \citep{lu2019vilbert} (36 Region Proposals) & 76.57   \\
    ViLBERT \citep{lu2019vilbert} (72 Region Proposals) & 79.39  \\
    ViLBERT \citep{lu2019vilbert} (108 Region Proposals) & 80.83 \\
    \bottomrule
  \end{tabular}
\end{table}



\subsection{Words matter more than grammar or semantics}
\label{experiments_ques_sem}

Next, we evaluated the influence of question semantics in driving the visual attention mechanism. Given a question/image pair, we randomly shuffled the order of words in the question and then forward propagated the question and the image through the ViLBERT model \citep{lu2019vilbert}. For instance, a question like 'What color is the floor?' could become 'Is color floor what the?'. The new question makes no semantic or grammatical sense. The shuffling procedure was done only at test time, while the model was trained with the words in the original order. 

\begin{figure}[htbp]
  \centering
  \includegraphics[width=8.98cm, height=6.7cm]{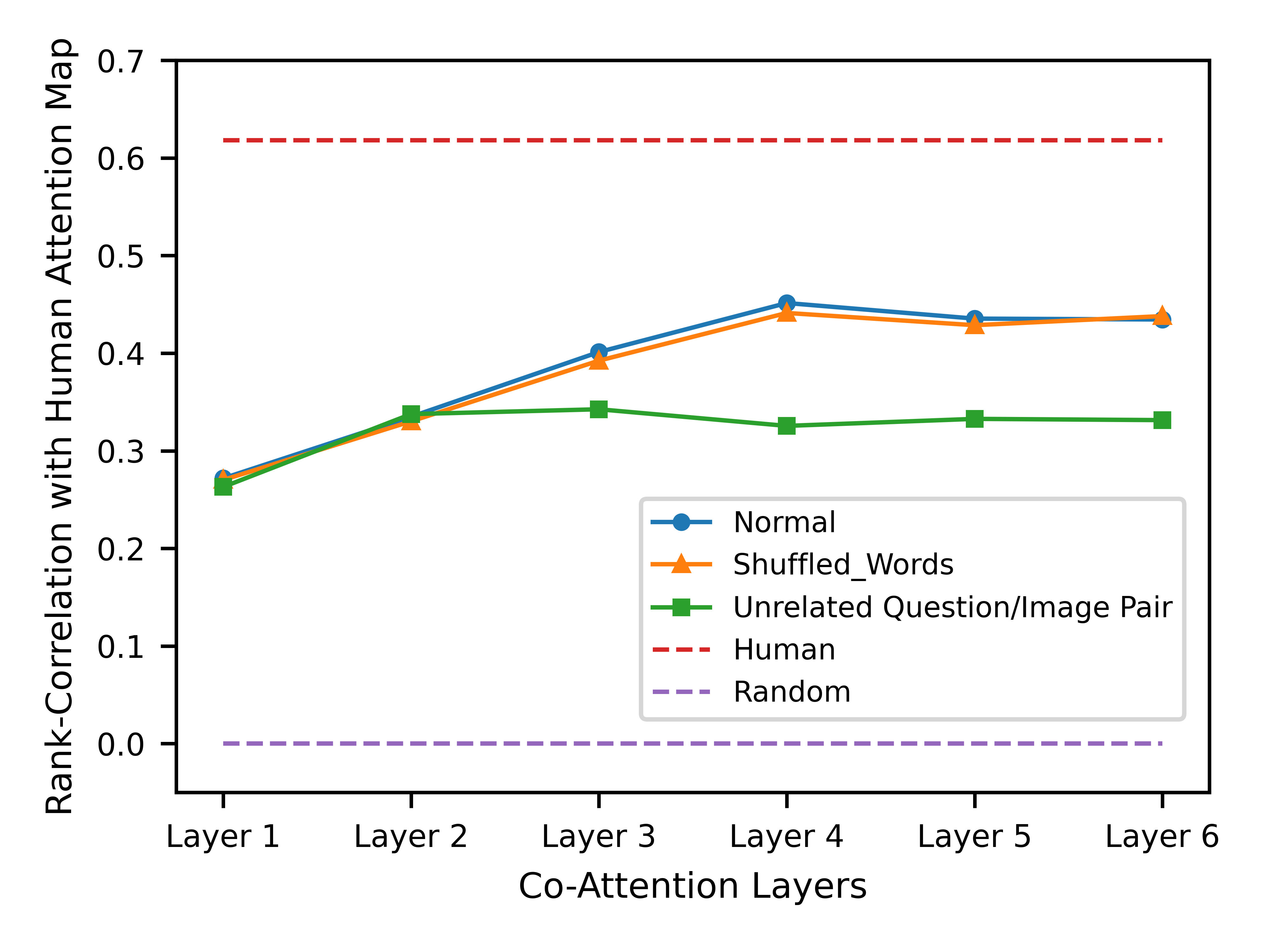}
  \caption{
  \textbf{The semantics of the question plays little role in driving the model's attention map.}
  Similarity between model and human attention maps ($\rho$, using 36 region proposals) for each of the 6 co-attention layers for the default (normal) model (blue), for the shuffled words condition (orange), and a condition where the image is paired with a random question (green). The format is similar to \textbf{Fig.~\ref{figure_experiments_roi}}, showing the between-human upper bound and the random levels.
  There is minimal change in $\rho$ after shuffling the words, indicating that semantics has little influence on ViLBERT's \citep{lu2019vilbert} attention. 
  }
  \label{figure_experiments_ques_sem}
\end{figure}
\begin{figure}[htbp]
  \centering

  \includegraphics[width=13.98cm, height=15cm]{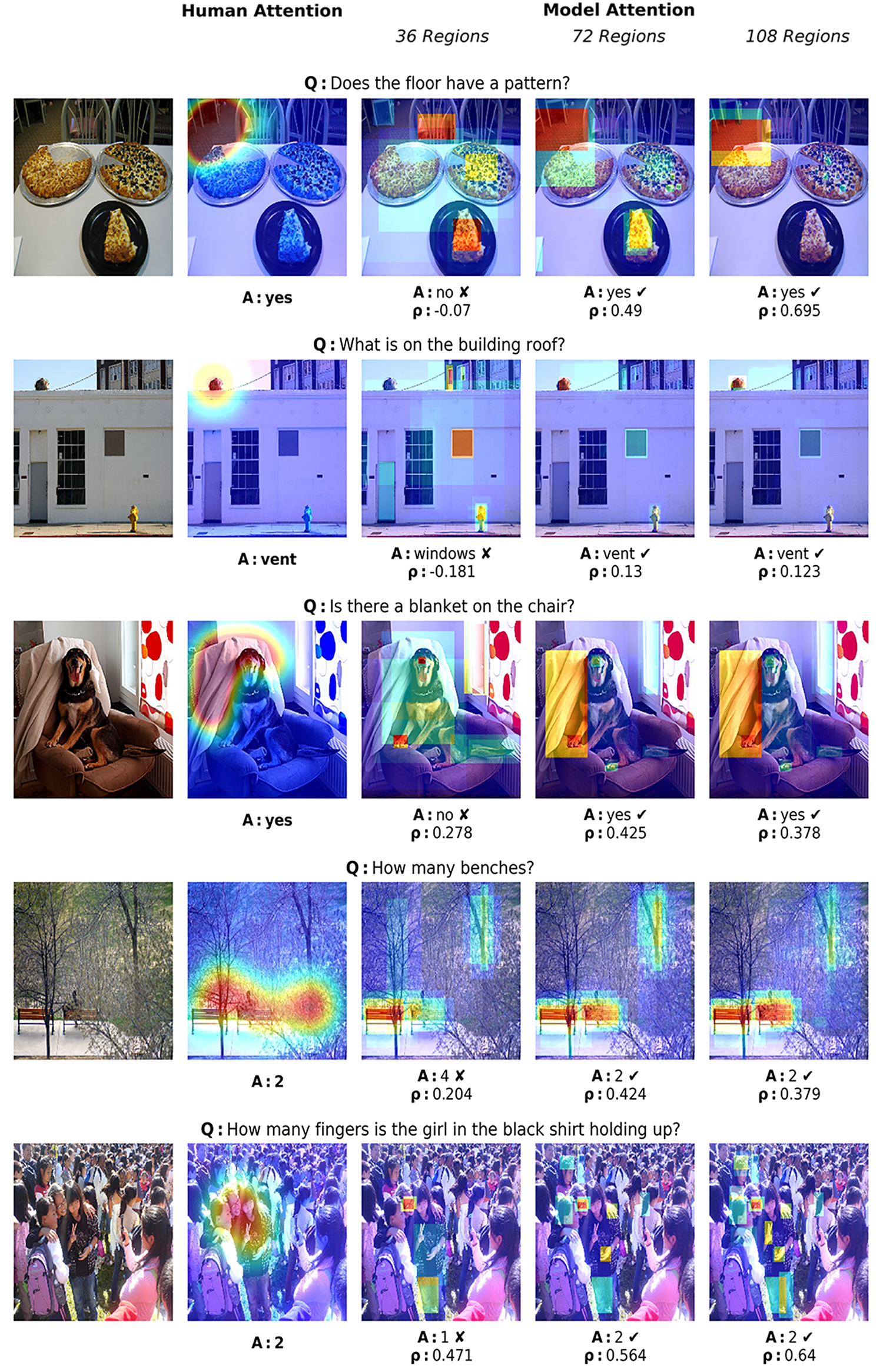}
  \includegraphics[scale=0.3]{./Images/colorbar.png}
  
  \caption{\textbf{Visualization for different question/image pairs and their corresponding attention maps across multiple controls}. Column 1 shows the input image, column 2 contains the human attention maps and Column 3, 4, and 5 show ViLBERT's \citep{lu2019vilbert} attention map for \textbf{Normal}, \textbf{Shuffled\_Words}, and \textbf{Unrelated Question/Image Pair} conditions, respectively. The answers in bold are ground-truth and the predicted answers are not in bold (see \textbf{Appendix A.2} for extended analyses).
  }
  \label{figure_qual_ques_sem}
\end{figure}

We expected that the rank-correlation of the model's attention with human attention for these modified questions should drop along with the VQA accuracy. However, the results did not match our expectations (\textbf{Fig.~\ref{figure_experiments_ques_sem}}, and visualization examples in \textbf{Fig.~\ref{figure_qual_ques_sem}}). There was only a minimal drop in the degree of similarity of the attention maps upon shuffling the word order. For example, in \textbf{Fig.~\ref{figure_qual_ques_sem}} row 1, ``What color is the floor?'' led to the correct answer (brown) and $\rho=0.548$ and the shuffled version ``Is color floor what the?'' also led to the correct answer and $\rho=0.556$. These results suggest that the question grammar and semantics play little to no role in modulating visual attention. Instead, the presence of specific keywords in the question is responsible for driving attention. 
Most of the visual grounding here is based on object-centric concepts rather than the overall semantics of the question.

The model's VQA accuracy dropped considerably after shuffling the words (\textbf{Table~\ref{acc_ques_sem_table}}). Thus, while attention seems to be largely independent of grammar and semantics, the ability to answer the questions correctly does require some notion of grammar and/or semantic information.

\begin{table}[htbp]
  \caption{VQA accuracy of ViLBERT \citep{lu2019vilbert} in different controls. Note that the reported accuracy is over question/image pairs in VQA-HAT \citep{vqahat} validation set. Refer section~\ref{experiments_ques_sem} for more details.\\}
  \label{acc_ques_sem_table}
  \centering
  \begin{tabular}{cc}
    \toprule
    Method     & VQA Accuracy      \\
    \midrule
    ViLBERT \citep{lu2019vilbert} (Normal) & 76.57   \\
    ViLBERT \citep{lu2019vilbert} (Shuffled Words) & 60.2  \\
    ViLBERT \citep{lu2019vilbert} (Unrelated Question/Image Pair) & 10.8   \\
    \bottomrule
  \end{tabular}
\end{table}

Given that attention was not dependent on the semantic content, we wondered whether it is possible that the model was focusing exclusively on visual information and simply ignoring the language part to drive attention allocation. To assess this possibility, we paired images with another randomly chosen question and compared the human attention maps with a given image/question pair and the model attention maps with the same image but a random question (\textbf{Fig.~\ref{figure_experiments_ques_sem}}). 
The rank-correlation in the case of Unrelated Question/Image Pair was largely driven by the visual input, any contribution from language in this case would be spurious. 

Following the example in \textbf{Fig.~\ref{figure_qual_ques_sem}}, row 1, the same image but using the question ``Is this singles or doubles?'' (instead of ``What color is the floor?''), led to the erroneous answer ``singles'' and $\rho=0.02$ (cf. $\rho=0.548$ for the correct question/image pair). The similarity with human attention was largely independent of the layer number but remained well above chance levels in the case of Unrelated Question/Image Pair (\textbf{Fig.~\ref{figure_experiments_ques_sem}}). Visual attention alone is sufficient to drive the rank-correlation with humans. Interestingly, even the unrelated question case shows higher similarity than previous benchmarks that combined visual and correct language information (\textbf{Table~\ref{models_rank_corr}}). For layers 3-6, the similarity with human attention dropped considerably with respect to the correct question condition. Thus, attention is largely dictated by visual information, combined with focused co-attention driven by the presence of specific key words irrespective of their ordering.





\subsection{Nouns drive attention}
\label{experiments_pos}
\begin{figure}[htbp]
  \centering
  \includegraphics[width=13.9cm, height=6.5cm]{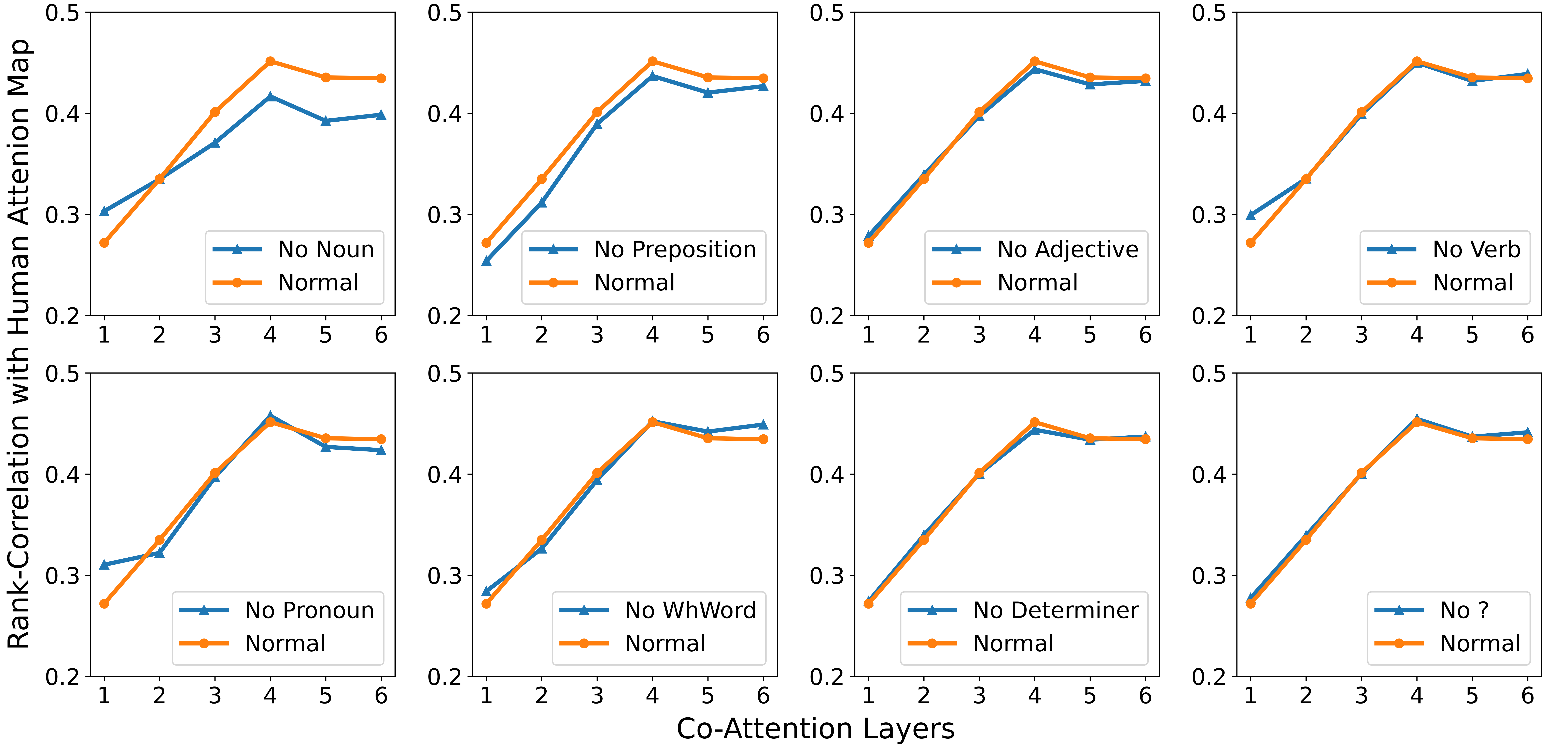}\\
  \caption{
  \textbf{Removing nouns, and to a lesser degree prepositions, led to a drop in similarity of attention maps}. Rank-correlation with human attention map ($\rho$) for each of the 6 co-attention layers upon removing different parts of speech (blue). The reduction in rank-correlation was maximal in the case of nouns, followed by prepositions and pronouns. Other parts of speech had little effect on the rank-correlation. Rank-correlation values shown here were averaged over question/image pairs containing words from the corresponding category (see Section~\ref{experiments_pos} for details). Error bars showing standard error of means are smaller than the symbol size in this plot.
  }
  \label{figure_experiments_pos}
\end{figure}

We quantified the importance of different parts of speech (POS) in guiding the model's attention to task-relevant image regions. Given a question and the corresponding image, we dropped words with a certain POS tag. For example, the question ``what is the girl holding?'' would become ``what is the holding?'' upon removing nouns. 
Then, we forward propagated the image and the modified question through the network and generated the corresponding attention maps, and computed the rank-correlation 
with the human attention maps. 
Similar to \citep{goyal2016towards}, we group POS tags into the following categories: Noun, Pronoun, Verb, Adjective, Preposition, Determiner, and Wh-Words. The Wh-Words category includes WP, WDT, and WRB tags containing words like who, which, and where respectively. We show the results of this experiment in \textbf{Fig.~\ref{figure_experiments_pos}}, using 36 region proposals. 

Consistent with our findings in Section~\ref{experiments_ques_sem} that words are more important than semantics, we noticed that nouns specifically played an important role in driving visual attention, followed by prepositions and pronouns. Given a question, nouns often help the model filter the relevant object categories from all the object region proposals. In addition, prepositions sometimes help guide attention based on spatial relationships between objects (see \textbf{Appendix A.3} for visualizations and additional qualitative results).

\subsection{Better performing VQA models show higher correlation with human attention maps}

In \textbf{Table~\ref{models_rank_corr}}, we show the VQA accuracy and rank-correlation of the model's attention maps and human attention maps for the following networks: ViLBERT \citep{lu2019vilbert}, Stacked Attention Network \citep{yang2016stacked} with 2 attention layers (SAN-2), Hierarchical Co-Attention Network \citep{lu2016hierarchical} with Word-Level (HieCoAtt-W), Phrase-Level (HieCoAtt-P), and Question-Level (HieCoAtt-Q). ViLBERT \citep{lu2019vilbert} uses a multi-modal transformer architecture while SAN-2 \citep{yang2016stacked} and HieCoAtt \citep{lu2016hierarchical} are based on CNN and LSTM architectures. 
The rank-correlation for the CNN/LSTM based models is considerably lower than the transformer-based model indicating a superior co-attention mechanism and better fusion of vision and language information in multi-modal transformers. Finally, it's interesting also to note that an increase in the VQA accuracy is accompanied by a better correlation with human attention.



\begin{table}[htbp]
  \caption{Accuracy for different VQA models on the VQA test-std set as reported in \citep{yang2016stacked, lu2016hierarchical, lu2019vilbert}. Error bars in rank-correlation here show standard error of means.\\}
  \label{models_rank_corr}
  \centering
  \begin{tabular}{ccc}
    \toprule
    Method   & Rank-Correlation  & VQA Accuracy      \\
    \midrule
    Random & 0.000 \textpm{} 0.001 & -   \\
    \cmidrule(lr){1-3}
    SAN-2 \citep{yang2016stacked} & 0.249 \textpm{} 0.004 & 58.9  \\
    \cmidrule(lr){1-3}
    HieCoAtt-W   \citep{lu2016hierarchical} & 0.246 \textpm{} 0.004 &   \\
    HieCoAtt-P \citep{lu2016hierarchical} & 0.256 \textpm{} 0.004 & 62.1  \\
    HieCoAtt-Q \citep{lu2016hierarchical} & 0.264 \textpm{} 0.004 &   \\
    \cmidrule(lr){1-3}
    ViLBERT \citep{lu2019vilbert} & \textbf{0.434 \textpm{} 0.006} & \textbf{70.92}        \\
    \cmidrule(lr){1-3}
    Human  & 0.618 \textpm{} 0.006 & -        \\
    \bottomrule
  \end{tabular}
\end{table}

\section{Conclusion \& Discussion}
\label{conclusion}

We conducted a series of experiments to interpret and study co-attention transformer layers and their role in aiding rich cross-modal interactions. We probed the modulation from language to vision in these co-attention layers and compared them with human attention maps. Transformer models lead to a substantial improvement in the similarity of attention maps with humans.
In addition, the attention maps of VQA models with higher accuracy are better correlated with human attention maps
Interestingly, the overall question semantics play a minimal role in guiding visual attention. Attention is governed by the visual inputs and by the presence of key nouns in the question.

The interpretability of multi-modal transformers has received little attention, despite their notable success in terms of performance metrics. While we are enthusiastic about recent advancements in Vision-Language models, it is also critical and instructive to examine transformer layers carefully. We illustrate through visualizations the observation that the object-based region proposals often act as a bottleneck and prevent the network from looking at task-relevant regions. There remains a large gap in accuracy between state-of-the-art VQA models and  human performance. At the same time, even though our results demonstrate that co-attention transformer layers yield a large boost to the congruency of attentional modulation in models and humans with respect to previous baselines, there is also a gap in the similarity of attention maps. We argue that this two gaps are related: building models that better capture human attention maps, perhaps by emphasizing the role of word combinations and semantics, can bring fundamental improvements in future VQA networks. 



\bibliographystyle{plainnat}
\bibliography{main}

\medskip

\newpage

\appendix

\section{Appendix}

\subsection{Additional qualitative results}

\vspace{0.1cm}

\begin{figure}[hbt!]
  \centering
  \includegraphics[width=11cm, height=17.5cm]{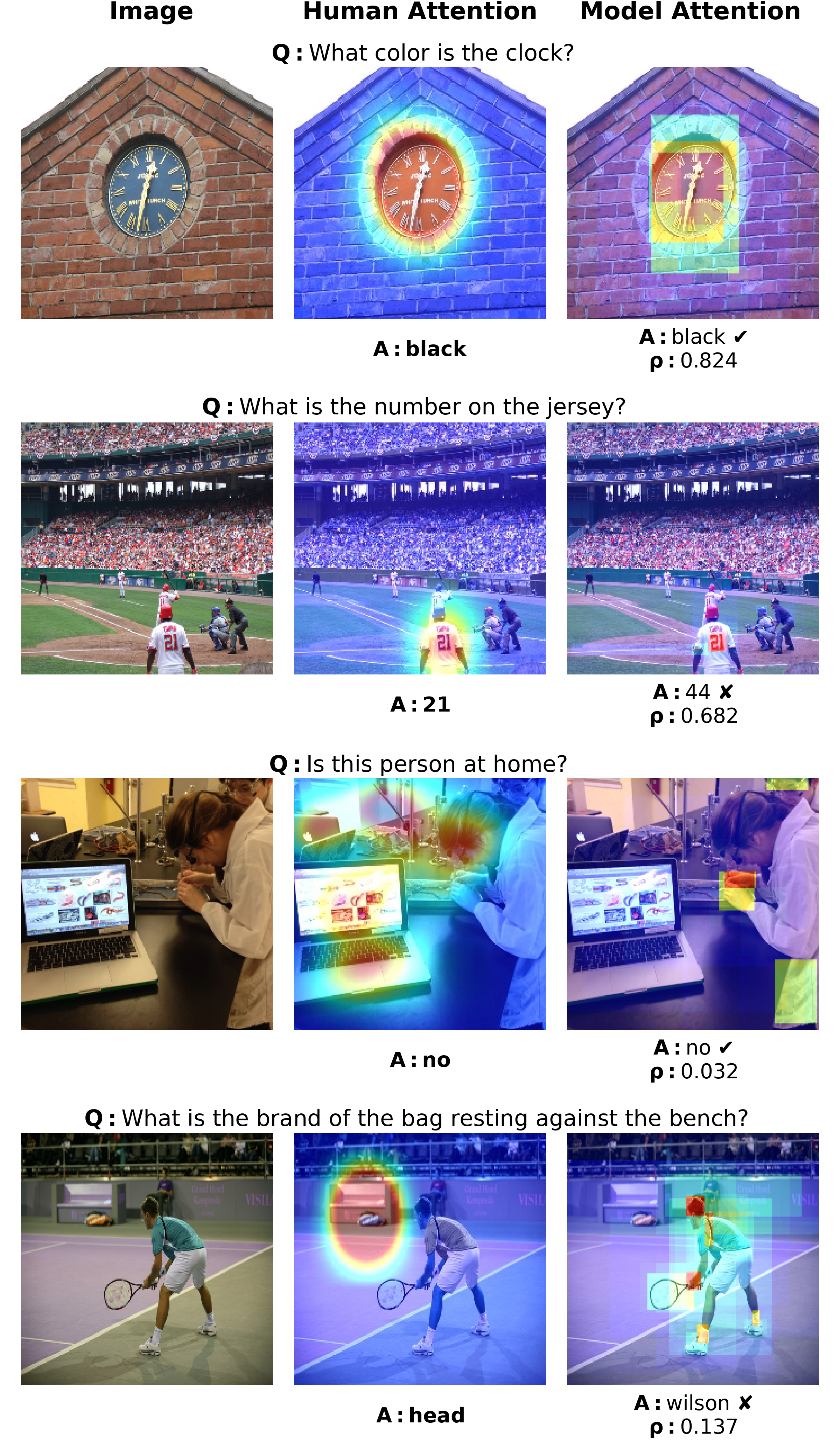}
  \includegraphics[scale=0.3]{./Images/colorbar.png}
  
  \caption{Row 1: high rank-correlation with 100\% accuracy, Row 2: high rank-correlation with 0\% accuracy, Row 3: low rank-correlation with 100\% accuracy, Row 4: low rank-correlation with 0\% accuracy. Column 1 shows the input image, column 2 contains the human attention maps, and column 3 shows ViLBERT's attention map. The answers in bold are ground-truth and the predicted answers are not in bold.
  }
\end{figure}

\subsection{Object region proposals act as a bottleneck}
  
  
  \begin{figure}[hbt!]
  \centering
  \includegraphics[width=13.98cm, height=18cm]{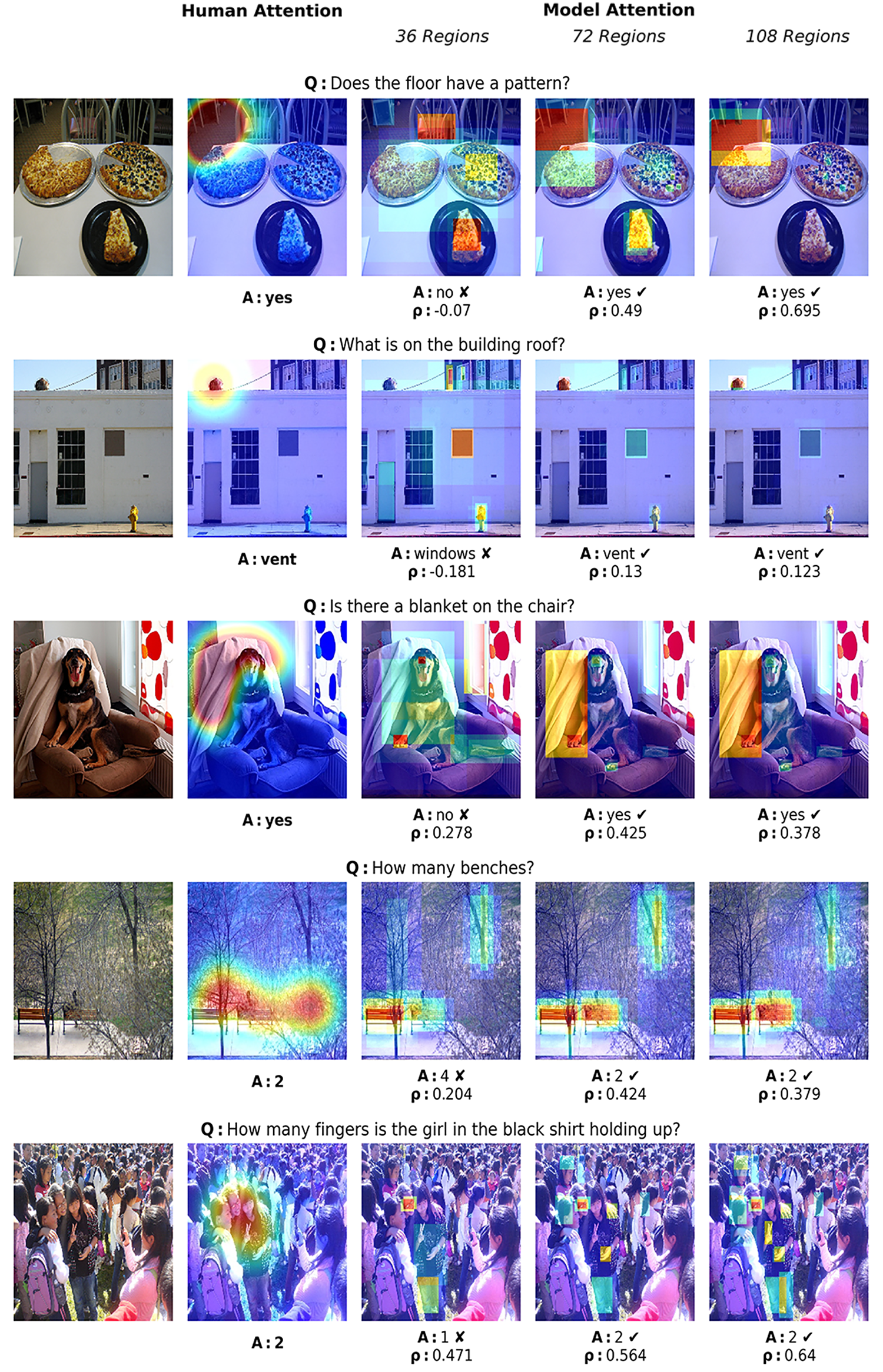}
  \includegraphics[scale=0.3]{./Images/colorbar.png}
  
  \caption{\textbf{Visualization for cases where number of regions proposals act as a bottleneck and restrict the network from attending to task-relevant regions.} 
  Column 1 shows the input image, column 2 contains the human attention maps, and Column 3,4, and 5 show ViLBERT's attention map for 36, 72, and 108 regions respectively. The answers in bold are ground-truth and the predicted answers are not in bold.
  }
  \end{figure}

\subsection{Question semantics play little role in visual attention}
  
  \begin{figure}[hbt!]
  \centering
  \includegraphics[width=13.98cm, height=18cm]{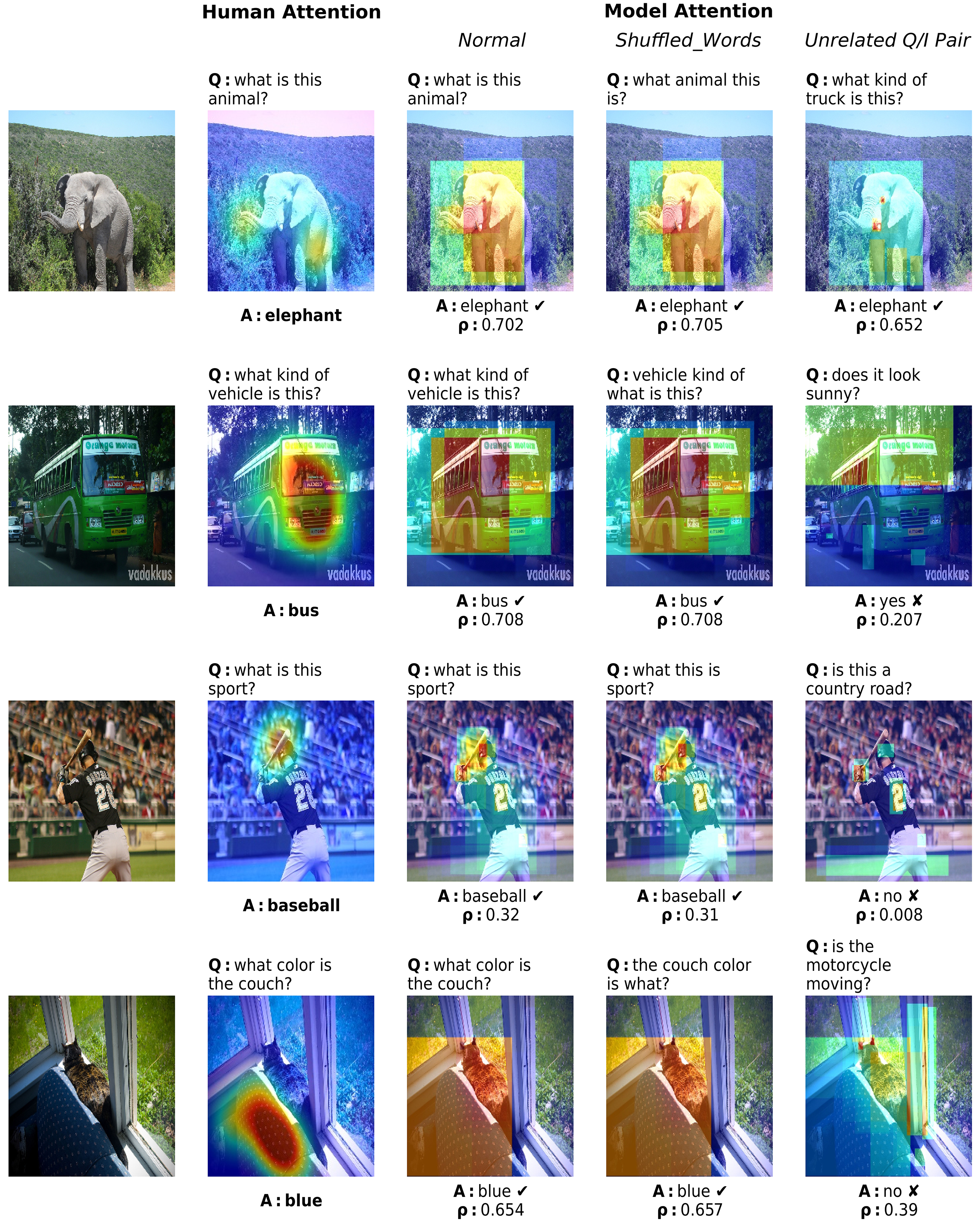}

  \includegraphics[scale=0.3]{./Images/colorbar.png}
  
  \caption{\textbf{Additional visualizations for different question/image pairs and their corresponding attention
  maps across multiple controls.} 
  Column 1 shows the input image, column 2 contains the human attention maps, and Column 3,4, and 5 show ViLBERT's attention map for \textbf{Normal}, \textbf{Shuffled\_Words}, and \textbf{Unrelated Question/Image Pair} conditions, respectively. The answers in bold are ground-truth and the predicted answers are not in bold.
  }
\end{figure}

\subsection{Importance of certain POS tags in guiding model's attention}

  \begin{figure}[hbt!]
  \centering
  \includegraphics[width=13.98cm, height=4.8cm]{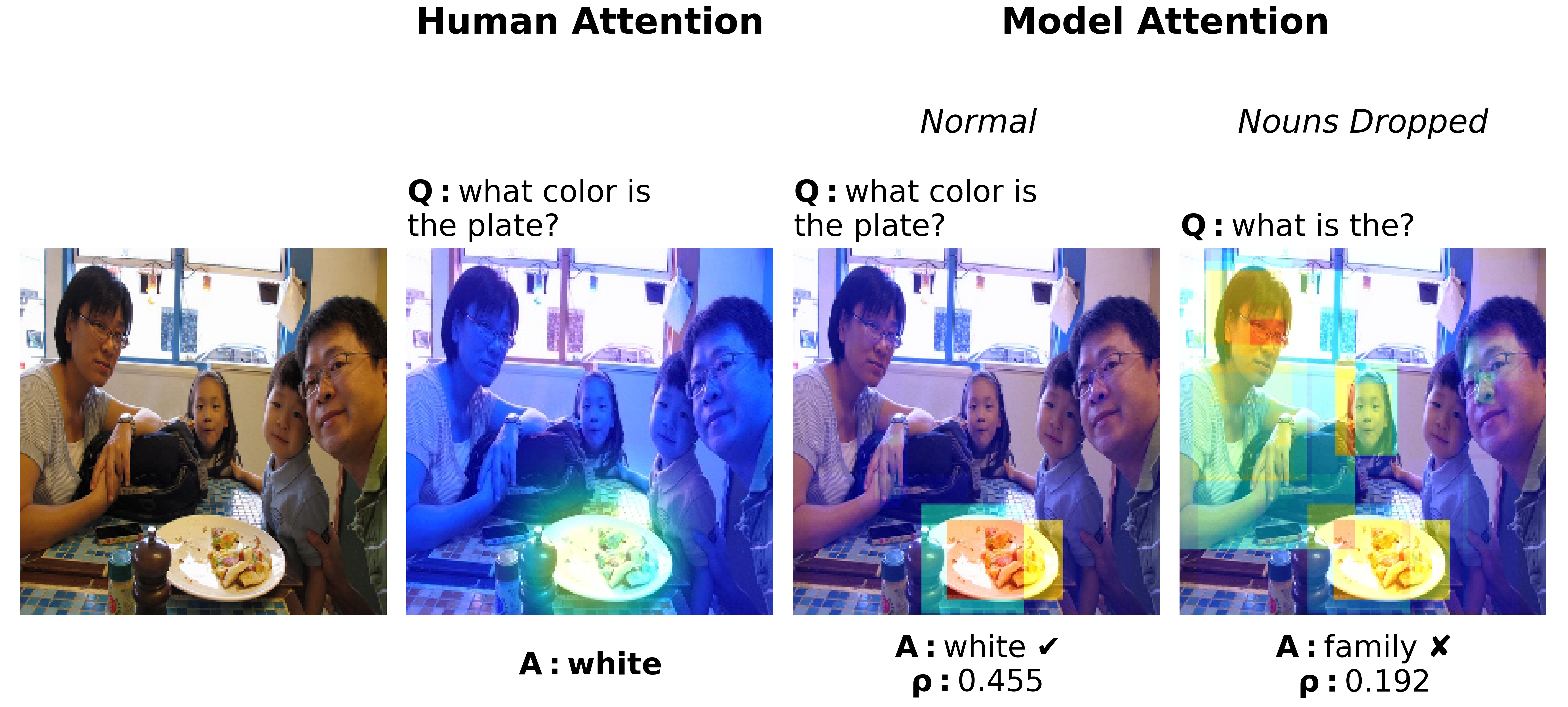}
  \includegraphics[width=13.98cm, height=4.6cm]{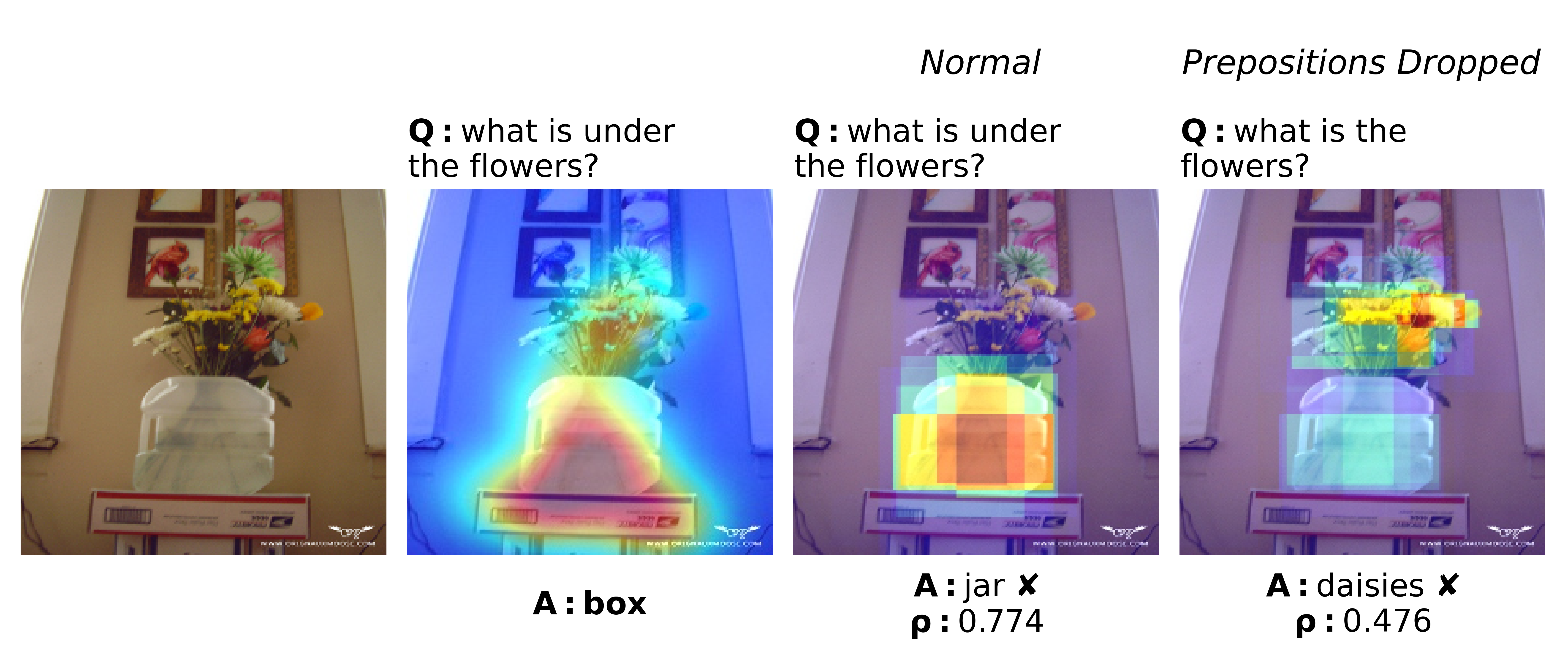}
  \includegraphics[width=13.98cm, height=4.6cm]{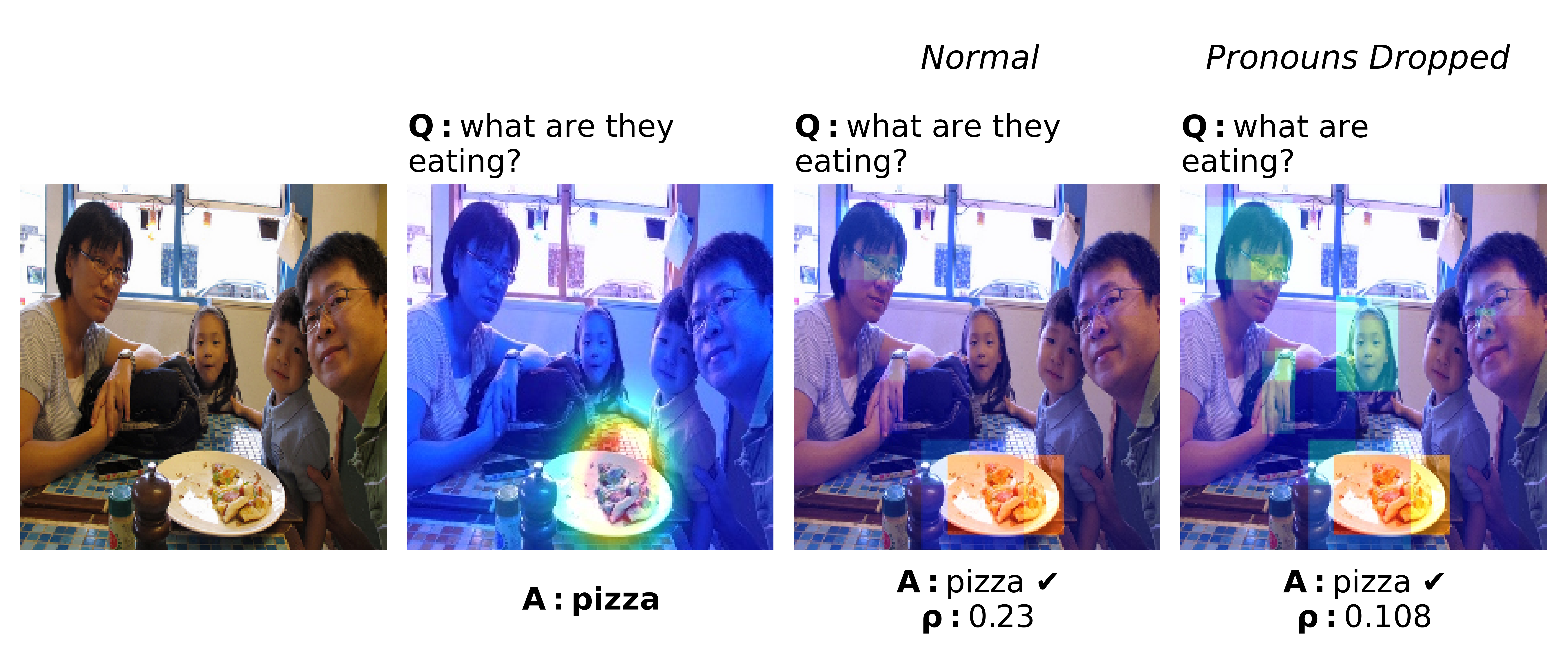}
  \includegraphics[width=13.98cm, height=4.6cm]{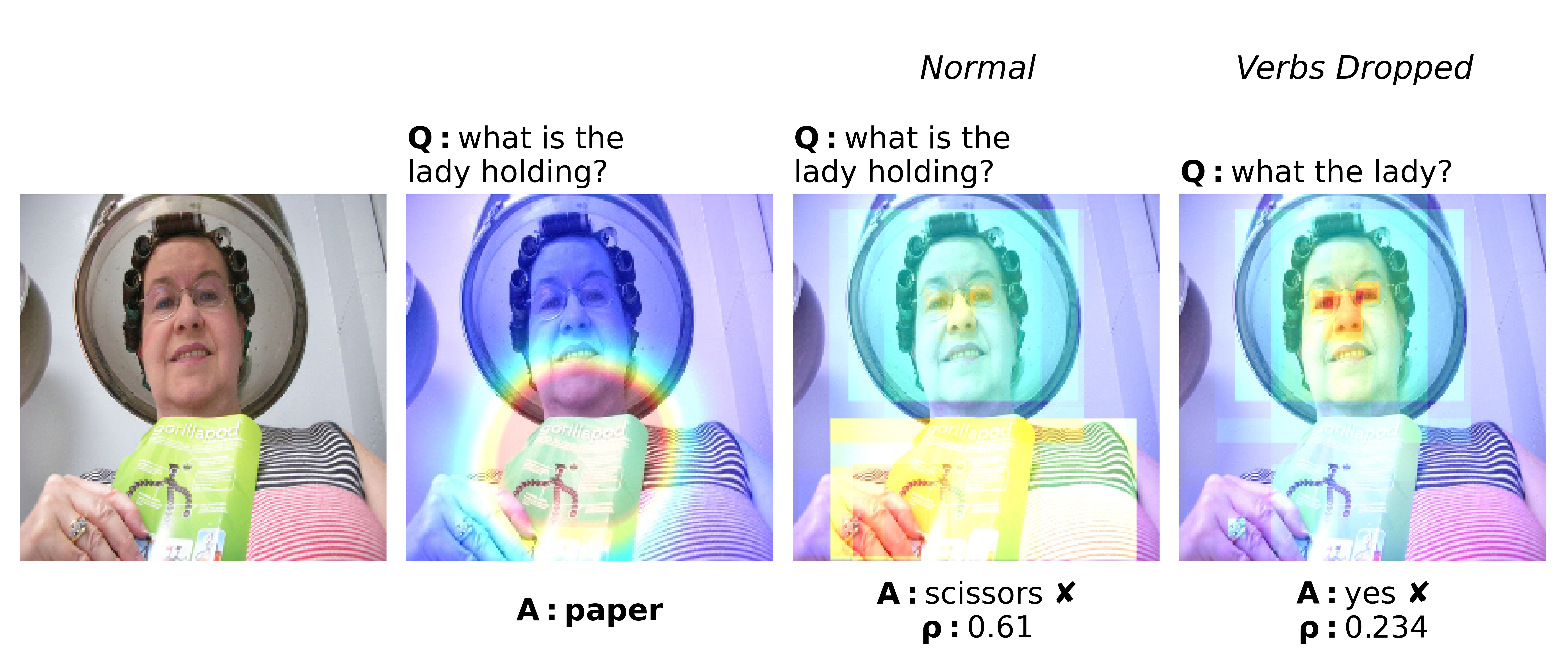}
  \includegraphics[scale=0.3]{./Images/colorbar.png}
  
  \caption{\textbf{Visualization for different question/image pairs and their corresponding attention
  maps after dropping words with certain POS tags.} Row 1: Nouns dropped, Row 2: Prepositions dropped, Row 3: Pronouns dropped, Row 4: Verbs dropped. The answers in bold are ground-truth and the predicted answers are not in bold.
  }
\end{figure}


\end{document}